# A generic system for critiquing physicians' prescriptions: usability, satisfaction and lessons learnt

Jean-Baptiste LAMY[a,1], Vahid EBRAHIMINIA[a], Brigitte SEROUSSI[a], Jacques BOUAUD[b], Christian SIMON[c], Madeleine FAVRE[d,e], Hector FALCOFF[d,e] and Alain VENOT[a]

[a] *Laboratoire d'Informatique Médicale et Bioinformatique (LIM&BIO), UFR SMBH, Université Paris 13, Bobigny, France*
[b] *AP-HP, DSI, STIM, Paris, France; INSERM, UMRS 872, eq. 20, Paris, France*
[c] *Silk Informatique, 40 bis avenue du général Patton, Angers, France*
[d] *Université Paris Descartes, Faculté de Médecine, Département de Médecine Générale, Paris*
[e] *Société de Formation Thérapeutique du Généraliste (SFTG), Paris, France*

**Abstract.** Clinical decision support systems have been developed to help physicians to take clinical guidelines into account during consultations. The ASTI critiquing module is one such systems; it provides the physician with automatic criticisms when a drug prescription does not follow the guidelines. It was initially developed for hypertension and type 2 diabetes, but is designed to be generic enough for application to all chronic diseases.

We present here the results of usability and satisfaction evaluations for the ASTI critiquing module, obtained with GPs for a newly implemented guideline concerning dyslipaemia, and we discuss the lessons learnt and the difficulties encountered when building a generic DSS for critiquing physicians' prescriptions.

## Introduction

Clinical guidelines (CG) provide physicians with recommendations, but paper guidelines are difficult to use effectively during medical consultations[1]. This difficulty has led to the development of decision support systems (DSS) based on CG[2]. The ASTI project aims to develop a DSS to help physicians to take into account the treatment recommendations expressed in CG for chronic diseases[3]. ASTI includes a critiquing module that is automatically activated when the physician writes a drug prescription, and which issues an alert if the prescription does not follow the CG. The critiquing module was initially developed for hypertension and type 2 diabetes. However, unlike many DSS which focus on a single CG, ASTI is designed to be generic enough to cover

---

all chronic diseases. To ensure the generic aspect of the system, a new CG concerning dyslipaemia[4] was implemented, and evaluated in the current study.

The critiquing module[5] and the validation of its knowledge bases[6] have been presented elsewhere. The CG recommendations are modeled in the critiquing module's knowledge base, and are then automatically translated into critiquing rules of the form "if physician prescribed treatment X to a patient with clinical condition P, then show criticism C". An inference engine applies these rules, and has been integrated into éO généraliste, an electronic patient record (EPR) for general practitioners (GPs). Drug prescriptions, laboratory test results and some clinical conditions are automatically extracted from the EPR and used by the critiquing module. Other clinical conditions required by the critiquing module are entered manually by physicians on a special form integrated in the EPR and displayed when a patient is included in the ASTI study at the beginning of the consultation.

We present here the results of the usability and satisfaction evaluations of the ASTI critiquing module for dyslipaemia, and we discuss the lessons learnt and the difficulties we encountered in the construction of this generic DSS.

**1. Methods**

A knowledge base has been designed and tested for the CG relating to dyslipaemia, as previously described[5,6]. It includes 28 decision criteria (patient's clinical conditions, laboratory test results, etc.), 15 drug treatments and 17 recommendations, resulting in 73 critiquing rules. We evaluated the critiquing module for dyslipaemia in the laboratory with 33 GPs. The GPs were éO users who volunteered to participate.

Two evaluations were performed with the ASTI critiquing module for dyslipaemia. Usability was evaluated using five simulated cases. These cases were derived, by an expert, from real cases, and were selected to cover the various aspects of the CG. The GPs were first briefly introduced to the use of the ASTI critiquing module. They were then asked, for each case, to code the data for the patient into the EPR and to enter two prescriptions: the usual prescription the doctor would write and a prescription that he or she did not consider to satisfy the CG. For each prescription, the physicians were asked to indicate whether they expected an alert, whether an alert was raised, whether the alert (or the absence of it) was justified, and whether the explanations and proposals accompanying the alert were appropriate. Additional textual comments were possible.

Satisfaction was evaluated just after the GPs had used the system. This evaluation was based on seven sentences. For each sentence, the GP had to tick one of four boxes, indicating strong agreement with the sentence, weak agreement, weak disagreement or strong disagreement. The evaluation was followed by a focus group, during which GPs were asked about the system, the way they used it and their feelings about it.

**2. Results**

The usability evaluation involved 299 prescriptions (less than 2x5x33, because some GPs did not reply to all questions, *e.g.* they rarely tried a second prescription if the first one was already criticized), divided as shown in Figure 1. The system's specificity was 94±1.4% (95% confidence interval) and the sensitivity 84±2.1%. The 136 true positives includes both prescriptions criticized as expected, or unexpectedly if the GP then agreed with the criticism; for 114 (84±2.1%) of the true positives, the GP considered the system's explanations and treatment proposals as appropriate. In 80±2.3% of cases, the system raised an appropriate criticism or was silent with good reason.

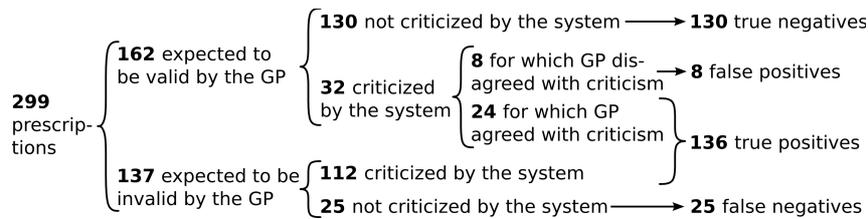

**Figure 1.** Results of the usability evaluation.

The results of the satisfaction evaluation are shown in Table 1. The physicians were interested in receiving automatic criticisms about their prescriptions and they found the ASTI critiquing module easy to use. However, they also felt that the use of the module interfered with doctor-patient relations. Further discussions with physicians showed that this problem was related to the time required to code the clinical context for the patient in the form displayed during the first consultation with the patient.

**Table 1.** Evaluation of satisfaction, expressed in percent (%)

| Question | Agreement | | Disagreement | |
|---|---|---|---|---|
| | Strong | Weak | Weak | Strong |
| I would like to receive automatic criticisms or suggestions relating to my prescriptions | 39 | 58 | 3 | 0 |
| The ASTI critiquing module is easy to use | 3 | 88 | 9 | 0 |
| The response time of the ASTI critiquing module is satisfactory | 73 | 24 | 3 | 0 |
| The ASTI critiquing module is ergonomic | 18 | 73 | 9 | 0 |
| The ASTI critiquing module can be effectively integrated into my daily practice | 28 | 66 | 6 | 0 |
| The ASTI critiquing module interferes little with my relationship with the patient | 0 | 27 | 67 | 6 |
| Extending the ASTI critiquing module to other guidelines would be a major step forward | 33 | 64 | 0 | 0 |

## 3. Discussion and conclusion

In this study, we evaluated the usability of and satisfaction with the ASTI critiquing module for dyslipaemia, a condition different from the hypertension and type 2 diabetes initially used for designing the system. Many other chronic diseases (*e.g.* asthma, cystic fibrosis,...) also have complex drug treatments evolving over long periods of time, the optimal treatment depending on several factors (lab test results, clinical conditions,...). The system's specificity was high, and the GPs expressed an interest in critiquing systems.

Similar evaluation designs, based on simulated cases, have already been used for evaluating DSS[7,8]. It would be interesting to carry out further evaluations of the critiquing system on real practice. Performing evaluation on voluntary GPs is a possible bias since they are usually enthusiastic, but this can hardly be avoided.

The first lesson learnt from this study is that it is possible to design a generic DSS supporting several CG for chronic diseases, despite the considerable heterogeneity of the various CG, which follow different treatment strategies (*e.g.* the CG for type 2 diabetes follows a "waterfall"-like linear strategy, depending on the stage of the disease, whereas the CG for dyslipaemia follow a "star"-like non-linear strategy depending on the type of dyslipaemia) and are often based on implicit knowledge (*e.g.* the CG does not generally mention that drug doses can be lowered to reduce adverse effects). Currently, five CG have been implemented: hypertension, type 2 diabetes, dyslipaemia, tobacco addiction, atrial fibrillation[5]. A few other DSS frameworks, such as Asbru[9] and others[10], have achieved a similar level of genericity.

Another lesson is that an automatic DSS, like this critiquing module, requires tight integration with the EPR used by the physician. However, as the various EPR include essentially the same patient data, it is possible to integrate a DSS into many different EPR. Semantic interoperability is easy to achieve, because a DSS usually has a limited number of decision criteria (*e.g.* 28 for dyslipaemia). During the ASTI project, the critiquing module was integrated into another EPR, ALMA Pro, produced by the ALMA association. The difficulties encountered during the integration process were organizational and financial rather than scientific or technical.

A third lesson is that physicians are interested in receiving automatic criticism on their prescriptions. This finding is consistent with other studies showing that physicians prefer automatic "background" DSS over "on-demand" DSS[11,12]. However, we also learnt that displaying the CG textual excerpts applying to the patient (as in older versions of the critiquing module, but not the one used during the evaluation) is not sufficient for the critiquing of physicians' prescriptions. Indeed, CG give recommendations such as "when the patient has clinical context C, drug W should be prescribed". However, they do not explain to the physician why the drug X, Y or Z he prescribed is not appropriate. For a given patient, many prescribing errors are possible and should receive different criticisms: *e.g.* drug X may be contraindicated due to another disease that the patient has, drug Y may be indicated only as a second-line treatment, and drug Z may already have been prescribed a year ago without success.

The major difficulty encountered is the coding of the patient's clinical conditions. The existing terminologies were developed for the coding of patient data in the EPR, but are not always relevant for coding decision criteria from CG. For instance, we were unable to code "family antecedent of myocardial infarction in the father before the age of 55" and "type 2 diabetes discovered at an advanced stage". Moreover, physicians do not usually code clinical elements in patient records. Instead, they tend to write them in free text, which is not usable by DSS as-is. While it might be possible to convince some physicians to code the principal diseases and antecedents of the patient, they are unlikely to code systematically complex decision criteria, such as those cited above. This problem has also been encountered in the ASTI guiding module[13], and is considered as one of the tenth "DSS grand challenges"[14]. By contrast, the coding of laboratory results and drug prescriptions is less problematic: test result criteria are generally simple in CG, and drug databases can be used for coding drug prescriptions.

Other difficulties relate to the CG themselves: they do not always provide clear recommendations, instead sometimes providing only "food for thought", which is not sufficient for critiquing. The various strength levels of recommendations are useful but not always mentioned in CG. In some situations, two CG may be contradictory. For example, the French CG for hypertension and for dyslipaemia give different formulas for determining cardiovascular risk level. Formalizing CG during their development may help to resolve these problems[15].

In conclusion, we have shown that the ASTI critiquing module, initially developed for hypertension and type 2 diabetes, is generic enough for application to dyslipaemia with good results. We have also shown that this module is of interest to physicians. The main difficulty is the coding of the patient's clinical conditions, but several approaches could be applied to this problem. First, graphical user interfaces or automated text processing tools could be designed to help physicians with data entry. Second, the coding of some clinical conditions could be done after the possible criticism rather than before, the physician explaining the reasons of his decision to the system. Finally, rather than executing the CG entirely, as the critiquing module does, other DSS might consist in presenting the CG to the physicians in a more usable form than plain text, possibly through graphical approaches.

**Acknowledgment**

We thank the HAS (*Haute Autorité de Santé*, the French health authority) and the CNAM (*Caisse Nationale d'Assurance Maladie*, the French health insurance fund for employees) for funding the ASTI project.